\documentclass[11pt]{article}
\usepackage{acl2016}
\usepackage{times}
\usepackage{latexsym}
\usepackage{amsmath}
\usepackage{amssymb}
\usepackage{graphicx}
\usepackage{ifthen}
\usepackage{booktabs}
\usepackage{caption}
\usepackage{multirow}
\usepackage{color}
\usepackage{subfig}
\usepackage{url}
\usepackage[utf8]{inputenc}

\newcommand\sZ{\ensuremath{\mathcal{Z}}}

\newcommand\bx{\ensuremath{\mathbf{x}}}

\newcommand\bz{\ensuremath{\mathbf{z}}}

\newcommand\FigTop[4]{\begin{figure}[t] \begin{center} \includegraphics[scale=#2]{#1} \end{center} \caption{\label{fig:#3} #4} \end{figure}}

      \newcommand\eqdef{\ensuremath{\stackrel{\rm def}{=}}}    \newcommand\refeqn[1]{(\ref{eqn:#1})}

\newcommand\refsec[1]{Section~\ref{sec:#1}}

\newcommand\reffig[1]{Figure~\ref{fig:#1}}

\newcommand\reftab[1]{Table~\ref{tab:#1}}

\ifthenelse{\isundefined{\definition}}{\newtheorem{definition}{Definition}}{}
\ifthenelse{\isundefined{\assumption}}{\newtheorem{assumption}{Assumption}}{}
\ifthenelse{\isundefined{\hypothesis}}{\newtheorem{hypothesis}{Hypothesis}}{}
\ifthenelse{\isundefined{\proposition}}{\newtheorem{proposition}{Proposition}}{}
\ifthenelse{\isundefined{\theorem}}{\newtheorem{theorem}{Theorem}}{}
\ifthenelse{\isundefined{\lemma}}{\newtheorem{lemma}{Lemma}}{}
\ifthenelse{\isundefined{\corollary}}{\newtheorem{corollary}{Corollary}}{}
\ifthenelse{\isundefined{\alg}}{\newtheorem{alg}{Algorithm}}{}
\ifthenelse{\isundefined{\example}}{\newtheorem{example}{Example}}{}
         
\newcommand\citep\cite
\newcommand\citet\newcite

\newcommand\nl[1]{``\emph{#1}''}
\newcommand\mwl[1]{\texttt{\footnotesize #1}}
\newcommand\wl[1]{\path{#1}}

\newcommand\alchemy{\textsc{Alchemy}}
\newcommand\scene{\textsc{Scene}}
\newcommand\tangrams{\textsc{Tangrams}}

\newcommand\exec{\text{Exec}}

\newcommand\za{z^\text{A}}
\newcommand\zb{z^\text{B}}
\newcommand\zc{z^\text{C}}
\newcommand\Piab{\Pi_{\text{A} \to \text{B}}}
\newcommand\Pibc{\Pi_{\text{B} \to \text{C}}}
 
\aclfinalcopy \def\aclpaperid{666} 
\title{Simpler Context-Dependent Logical Forms via Model Projections}

\author{
  Reginald Long \\
  Stanford University \\
  {\small \tt{reggylong@cs.stanford.edu}}
\And
  Panupong Pasupat \\
  Stanford University \\
  {\small \tt{ppasupat@cs.stanford.edu}}
\And
  Percy Liang \\
  Stanford University \\
  {\small \tt{pliang@cs.stanford.edu}}
}

\date{}

\begin{document}

\maketitle

\begin{abstract}

We consider the task of learning a context-dependent mapping
from utterances to denotations.
With only denotations at training time,
we must search over a combinatorially large space of logical forms,
which is even larger with context-dependent utterances.
To cope with this challenge,
we perform successive projections of the full model
onto simpler models that operate over equivalence classes of logical forms.
Though less expressive, we find that these simpler models are much faster
and can be surprisingly effective.
Moreover, they can be used to bootstrap the full model.
Finally, we collected three new context-dependent semantic parsing datasets,
and develop a new left-to-right parser.
 \end{abstract}

\section{Introduction}

Suppose we are only told that a piece of text (a command)
in some context (state of the world)
has some denotation (the effect of the command)---see \reffig{runningExample} for an example.
How can we build a system to learn from examples like these
with no initial knowledge about what any of the words mean?

We start with the classic paradigm of training semantic parsers
that map utterances to logical forms,
which are executed to produce the denotation
\cite{zelle96geoquery,zettlemoyer05ccg,wong07synchronous,zettlemoyer09context,kwiatkowski10ccg}.
More recent work learns directly from denotations
\cite{clarke10world,liang2013lambdadcs,berant2013freebase,artzi2013weakly},
but in this setting,
a constant struggle is to contain the exponential explosion of possible logical forms.
With no initial lexicon and longer context-dependent texts,
our situation is exacerbated.

\FigTop{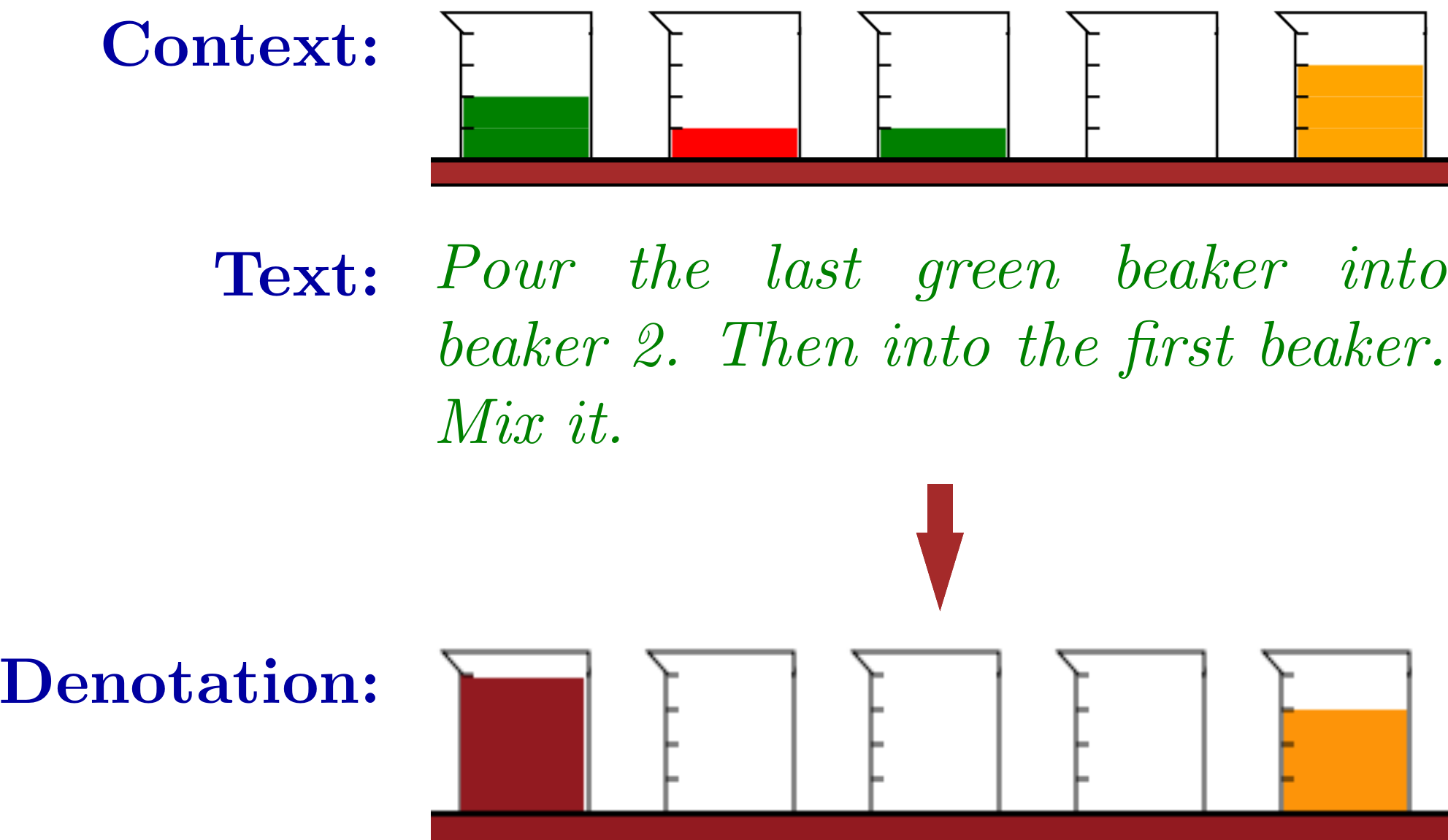}{0.3}{runningExample}{
  Our task is to learn to map a piece of text in some context to a denotation.
  An example from the \alchemy{} dataset is shown.
  In this paper, we ask: what intermediate logical form is suitable
  for modeling this mapping?
        }

\begin{figure}[t]
\subfloat{\includegraphics[scale=0.33]{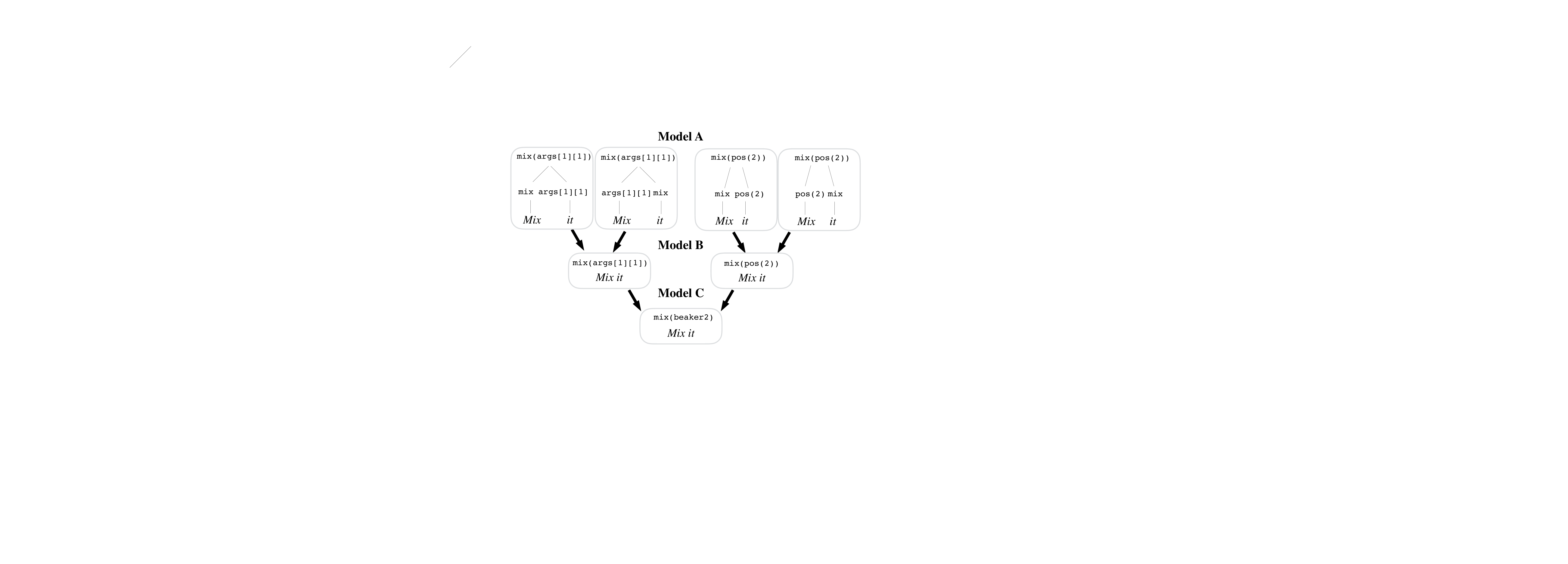}}
\caption{
Derivations generated for the last utterance 
in \reffig{runningExample}.
All derivations above execute to \wl{mix(beaker2)}.
Model A generates anchored logical forms (derivations) where words are aligned to predicates,
which leads to multiple derivations with the same logical form.
Model B discards these alignments, and
Model C collapses the arguments of the logical forms to denotations.
}
\label{fig:relaxExample}
\end{figure}

In this paper, we propose \emph{projecting} a full semantic parsing model onto
simpler models over equivalence classes of logical form derivations. As illustrated in \reffig{relaxExample},
we consider the following sequence of models:
\begin{itemize}
  \item \textbf{Model A}: our full model that derives logical forms
    (e.g., in \reffig{runningExample},       the last utterance maps to \wl{mix(args[1][1])})
    compositionally from the text so that
    spans of the utterance (e.g., \nl{it})     align to parts of the logical form (e.g., \wl{args[1][1]}, which retrieves an argument from a previous logical form).     This is based on standard semantic parsing (e.g., \citet{zettlemoyer05ccg}).
  \item \textbf{Model B}: collapse all derivations with the same logical form;
    we map utterances to full logical forms, but without an alignment
    between the utterance and logical forms.
    This ``floating'' approach was used in \citet{pasupat2015compositional} and \citet{wang2015overnight}.
  \item \textbf{Model C}: further collapse all logical forms whose top-level arguments have the same denotation.
    In other words, we map utterances to flat logical forms (e.g., \wl{mix(beaker2)}),     where the arguments of the top-level predicate are objects in the world.
    This model is in the spirit of \citet{yao2014freebase} and \citet{bordes2014qa},
    who directly predicted concrete paths in a knowledge graph for question answering.
\end{itemize}
Model A excels at credit assignment:
the latent derivation explains how parts of the logical form are triggered by
parts of the utterance.  The price is an unmanageably large search space, given
that we do not have a seed lexicon.
At the other end, Model C only considers a small set of logical forms,
but the mapping from text to the correct logical form is more complex
and harder to model.

We collected three new context-dependent semantic
parsing datasets using Amazon Mechanical Turk:
\alchemy{} (\reffig{runningExample}), \scene{} (\reffig{people}), and \tangrams{} (\reffig{tangrams}).
Along the way,
we develop a new parser which processes utterances
left-to-right but can construct logical forms without an explicit alignment.

Our empirical findings are as follows:
First, Model C is surprisingly effective,
mostly surpassing the other two given bounded computational resources (a fixed beam size).
Second, on a synthetic dataset, with infinite beam,
Model A outperforms the other two models.
Third, we can bootstrap up to Model A from the projected models with finite beam.

 \section{Task}

In this section,
we formalize the task and describe the new datasets we created for the task.

\subsection{Setup}

First, we will define the context-dependent semantic parsing task.
Define $w_0$ as the initial world state, which consists of a set of entities
(beakers in \alchemy{}) and properties (location, color(s), and amount filled).
The text $\bx$ is a sequence of utterances $x_1, \dots, x_L$.
For each utterance $x_i$ (e.g., \nl{mix}), we have a latent logical form $z_i$
(e.g., \wl{mix(args[1][2])}).
Define the context $c_i = (w_0, z_{1:i-1})$ to include the initial world state $w_0$ and the history
of past logical forms $z_{1:i-1}$.
Each logical form $z_i$ is executed on the context $c_i$
to produce the next state: $w_i = \exec(c_i, z_i)$ for each $i = 1, \dots, L$.
Overloading notation, we write $w_L = \exec(w_0, \bz)$, where $\bz = (z_1, \dots, z_L)$.

The learning problem is:
given a set of training examples $\{(w_0, \bx, w_L)\}$,
learn a mapping from the text $\bx$ to logical forms $\bz = (z_1, \dots, z_L)$
that produces the correct final state ($w_L = \exec(w_0, \bz)$).

\FigTop{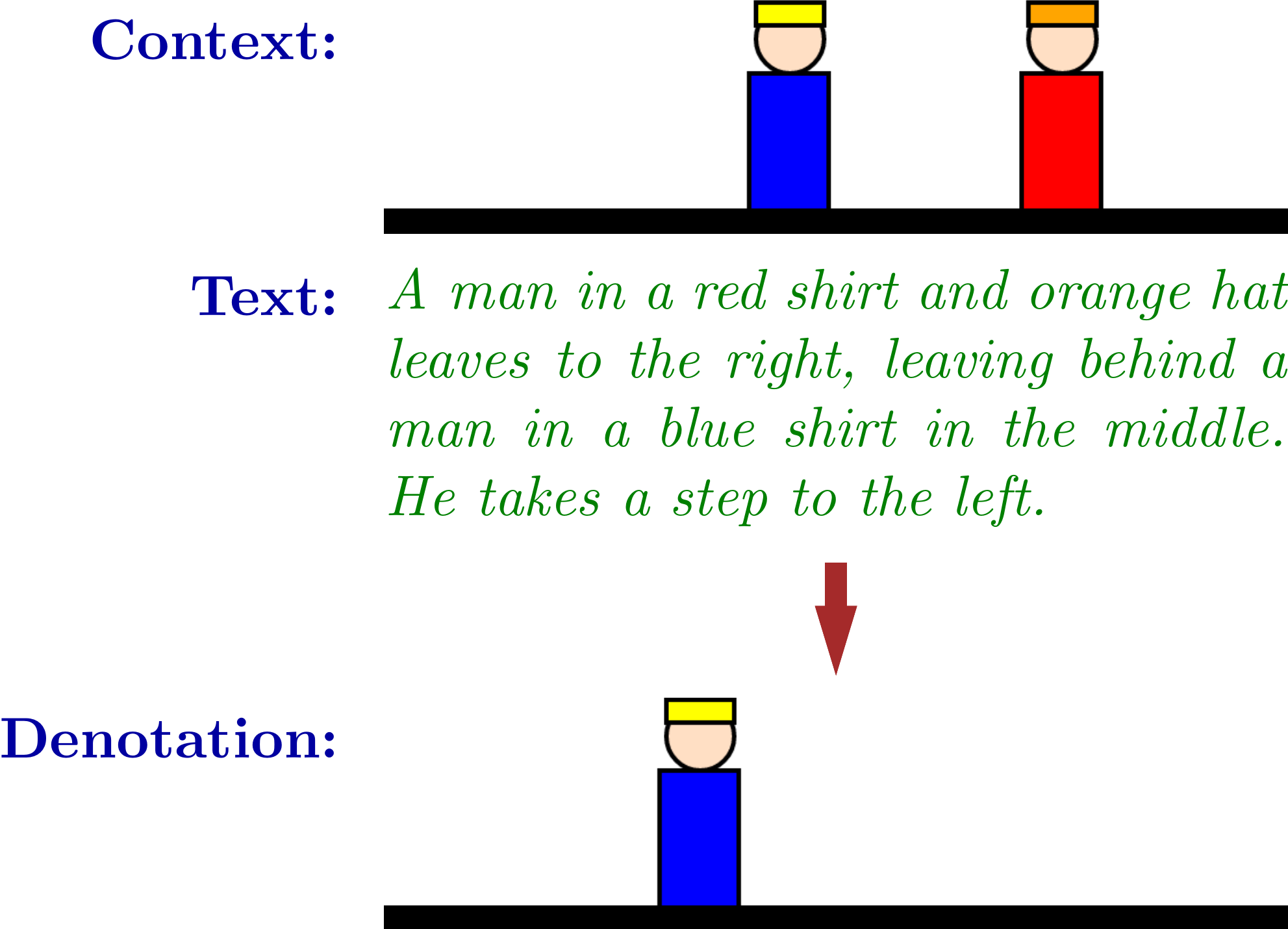}{0.3}{people}{\scene{} dataset:
Each person has a shirt of some color and a hat of some color.
They enter, leave, move around on a stage, and trade hats.
}
\FigTop{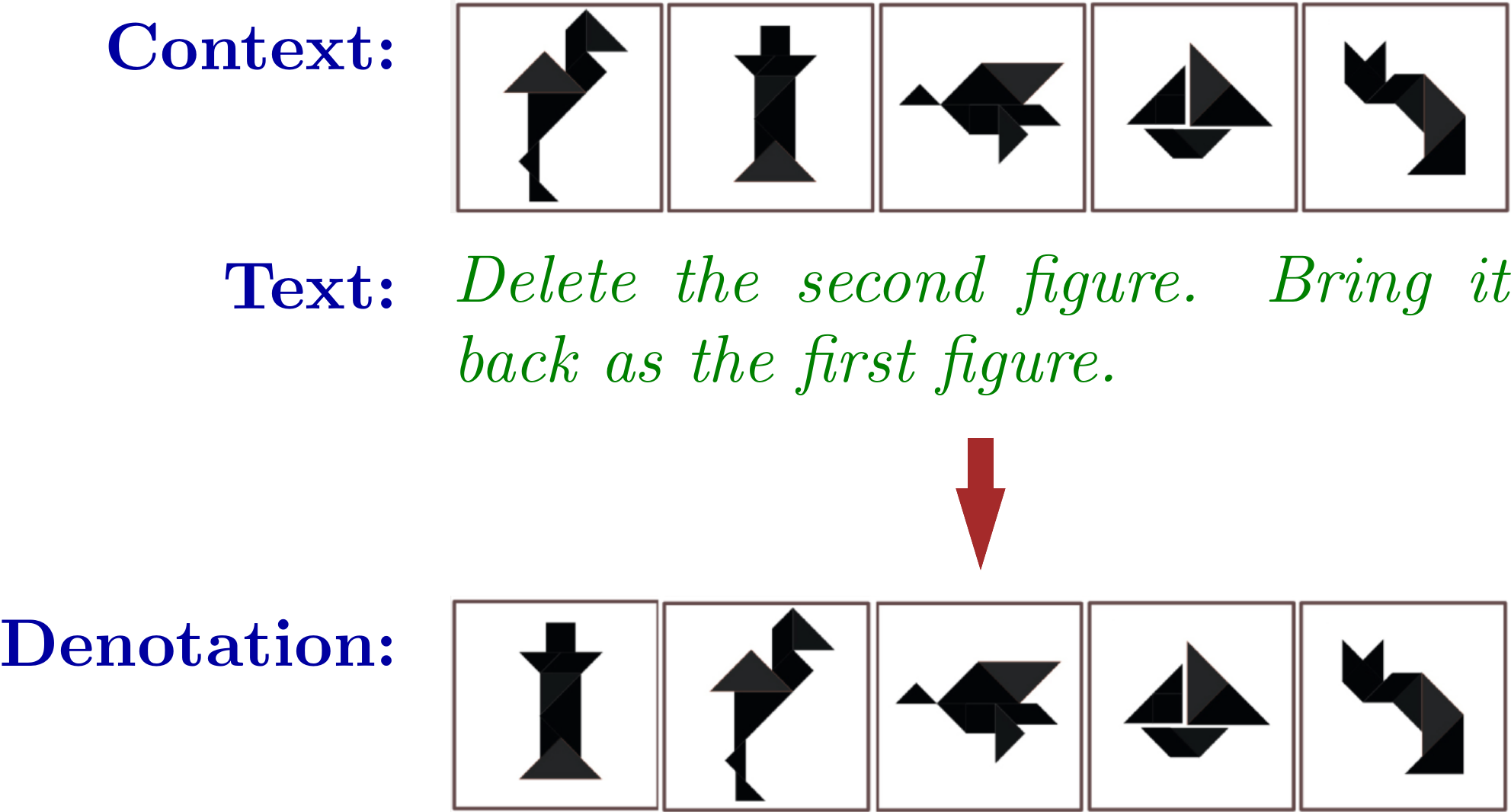}{0.3}{tangrams}{\tangrams{} dataset:
One can add figures, remove figures, and swap the position of figures.  All the figures slide to the left.
}

\subsection{Datasets}
\label{sec:datasets}

\begin{table*}[ht]
  \small
  \centering
\begin{tabular}{r|rrrrl}
  Dataset   & \# examples & \# train & \# test & words/example   & utterances \\ \hline
  \scene    & 4402       &  3363    & 1039    & 56.2             & \nl{then one more}, \nl{he moves back}       \\
  \alchemy  & 4560       &  3661    & 899     & 39.9             & \nl{mix}, \nl{throw the rest out}       \\
  \tangrams & 4989       &  4189    & 800     & 27.2             & \nl{undo}, \nl{replace it}, \nl{take it away}       \\
\end{tabular}
\caption{\label{tab:datasets}
We collected three datasets.
The number of examples, train/test split, number of tokens per example,
along with interesting phenomena are shown for each dataset.
}
\end{table*}

We created three new context-dependent datasets, \alchemy{}, \scene{}, and \tangrams{}
(see \reftab{datasets} for a summary),
which aim to capture a diverse set of context-dependent linguistic phenomena such as
ellipsis (e.g., \nl{mix} in \alchemy{}), anaphora on entities (e.g., \nl{he} in \scene{}),
and anaphora on actions (e.g., \nl{repeat step 3}, \nl{bring it back} in \tangrams{}).

For each dataset, we have a set of properties and actions.
In \alchemy{}, properties are \wl{color}, and \wl{amount};
actions are \wl{pour}, \wl{drain}, and \wl{mix}.
In \scene{}, properties are \wl{hat-color} and \wl{shirt-color};
actions are \wl{enter}, \wl{leave}, \wl{move}, and \wl{trade-hats}.
In \tangrams{}, there is one property (\wl{shape}),
and actions are \wl{add}, \wl{remove}, and \wl{swap}.
In addition, we include the position property (\wl{pos}) in each dataset.
Each example has $L=5$ utterances, each denoting some transformation of the world state.

Our datasets are unique in that they are grounded to a world state
and have rich linguistic context-dependence.
In the context-dependent ATIS dataset \citep{dahl1994expanding} used
by \citet{zettlemoyer09context},
logical forms of utterances depend on previous logical forms,
though there is no world state and the linguistic phenomena is limited to nominal references.
In the map navigation dataset \citep{chen11navigate},
used by \citet{artzi2013weakly},
utterances only reference the current world state.
\citet{vlachos2014new} released a corpus of annotated dialogues, which
has interesting linguistic context-dependence, but there is no world state.

\paragraph{Data collection.}

Our strategy was to automatically generate sequences of world states and ask Amazon
Mechanical Turk (AMT) workers to describe the successive transformations.
Specifically, we started with a random world state $w_0$.
For each $i = 1, \dots, L$, we sample
a valid action and argument (e.g., \wl{pour(beaker1,beaker2)}).
To encourage context-dependent descriptions,
we upweight recently used actions and arguments
(e.g., the next action is more like to be \wl{drain(beaker2)} rather than \wl{drain(beaker5)}).
Next, we presented an AMT worker with states $w_0, \dots, w_L$
and asked the worker to write a description in between each pair of successive
states.

In initial experiments,
we found it rather non-trivial to obtain interesting linguistic
context-dependence in these micro-domains:
often a context-independent utterance such as \nl{beaker 2}
is just clearer and not much longer than a possibly ambiguous \nl{it}.
We modified the domains to encourage more context.
For example, in \scene{}, we removed any visual indication of absolute position
and allowed people to only move next to other people.
This way, workers would say \nl{to the left of the man in the red hat}
rather than \nl{to position 2}.
 \section{Model}

\begin{table*}
  \small   \begin{center}
\begin{tabular}{lll}
  \toprule
  Property[$p$] Value[$v$]              & $\Rightarrow$ Set[$p(v)$]                          & all entities whose property $p$ is $v$ \\
  Set[$s$] Property[$p$]                & $\Rightarrow$ Value[$\text{argmin/argmax}(s, p)$]  & element in $s$ with smallest/largest $p$ \\
  Set[$s$] Int[$i$]                     & $\Rightarrow$ Value[$s[i]$]                        & $i$-th element of $s$ \\
  Action[$a$] Value[$v_1$] Value[$v_2$] & $\Rightarrow$ Root[$a(v_1, v_2)$]                  & top-level action applied to arguments $v_1,v_2$ \\
  \bottomrule
\end{tabular}
\caption{
\label{tab:grammar}
Grammar that defines the space of candidate logical forms.
Values include numbers, colors, as well as special tokens $\texttt{args}[i][j]$
(for all $i \in \{1, \dots, L\}$ and $j \in \{1,2\}$) that refer
to the $j$-th argument used in the $i$-th logical form.
Actions include the fixed domain-specific set
plus special tokens $\texttt{actions}[i]$ (for all $i \in \{1, \dots, L\}$),
which refers to the $i$-th action in the context.
}
\end{center}
\end{table*}
 \begin{table*}[t]
  \small
\begin{tabular}{lll}
 & Derivation condition & Example \\ 
\toprule
(F1) & $z_i$ contains predicate $r$ 					        & ($z_i$ contains predicate \wl{pour}, \nl{pour}) \\ 
(F2) & property $p$ of $z_i.b_j$ is $y$                         & (\wl{color} of arg 1 is \wl{green}, \nl{green}) \\ 
(F3) & action $z_i.a$ is $a$ and property $p$ of $z_i.y_j$ is $y$                         & (action is \wl{pour} and \wl{pos} of arg 2 is \wl{2}, \nl{pour, 2}) \\
(F4) & properties $p$ of $z_i.v_1$ is $y$ and $p'$ of $z_i.v_2$ is $y'$         & (\wl{color} of arg 1 is \wl{green} and \wl{pos} of arg 2 is \wl{2}, \nl{first green, 2}) \\
(F5) & arg $z_i.v_j$ is one of $z_{i-1}$'s args                          & (arg reused, \nl{it}) \\
(F6) & action $z_i.a = z_{i-1}.a$                                                  & (action reused, \nl{pour}) \\
(F7) & properties $p$ of $z_i.y_j$ is $y$ and $p'$ of $z_{i-1}.y_k$ is $y'$         & (\wl{pos} of arg 1 is \wl{2} and \wl{pos} of prev. arg 2 is \wl{2}, \nl{then}) \\
(F8) & $t_1 < s_2$           													   & spans don't overlap \\ 

\bottomrule
\end{tabular}
\caption{
  Features $\phi(x_i, c_i, z_i)$ for Model A:
  The left hand side describes conditions under which the system fires indicator features,
  and right hand side shows sample features for each condition.
  For each derivation condition (F1)--(F7),
  we conjoin the condition with the span of the utterance that the referenced actions and arguments align to.
  For condition (F8), we just fire the indicator by itself.
}
\label{tab:features}
\end{table*}
 
We now describe Model A, our full context-dependent semantic parsing model.
First, let $\sZ$ denote the set of candidate logical forms
(e.g., \wl{pour(color(green),color(red))}).
Each logical form consists of a top-level action with arguments,
which are either primitive values (\wl{green}, \wl{3}, etc.),
or composed via selection and superlative operations.
See \reftab{grammar} for a full description.
One notable feature of the logical forms is the context dependency:
for example, given some context $(w_0, z_{1:4})$,
the predicate \wl{actions[2]} refers to the action of $z_2$ and
\wl{args[2][1]} refers to first argument of $z_2$.\footnote{
These special predicates play the role of references in \citet{zettlemoyer09context}.
They perform context-independent parsing and resolve references,
whereas we resolve them jointly while parsing.}

We use the term \emph{anchored logical forms} (a.k.a. derivations) to refer
to logical forms augmented with alignments between sub-logical forms of $z_i$
and spans of the utterance $x_i$.
In the example above, \wl{color(green)} might align with \nl{green beaker} from \reffig{runningExample};
see \reffig{relaxExample} for another example.

\paragraph{Log-linear model.}

We place a conditional distribution over anchored logical forms $z_i \in \sZ$
given an utterance $x_i$ and context $c_i = (w_0, z_{1:i-1})$,
which consists of the initial world state $w_0$ and the history of past logical forms $z_{1:i-1}$.
We use a standard log-linear model:
\begin{align}
\label{eqn:model}
p_\theta(z_i \mid x_i, c_i) \propto \exp(\phi(x_i, c_i, z_i) \cdot \theta),
\end{align}
where $\phi$ is the feature mapping
and $\theta$ is the parameter vector (to be learned).
Chaining these distributions together,
we get a distribution over a sequence of logical forms $\bz = (z_1, \dots, z_L)$
given the whole text $\bx$:
\begin{align}
  p_\theta(\bz \mid \bx, w_0) = \prod_{i=1}^L p_\theta(z_i \mid x_i, (w_0, z_{1:i-1})).
\end{align}

\paragraph{Features.}

Our feature mapping $\phi$ consists of two types of indicators:
\begin{enumerate}
\item For each derivation, we fire features based on the structure of the logical form/spans.
\item For each span $s$ (e.g., \nl{green beaker})
aligned to a sub-logical form $z$ (e.g., \wl{color(green)}),
we fire features on unigrams, bigrams, and trigrams inside $s$
conjoined with various conditions of $z$.
\end{enumerate}
The exact features given in \reftab{features},
references the first two utterances of \reffig{runningExample}
and the associated logical forms below:
\begin{align*} 
  x_1 &= \text{\nl{Pour the last green beaker into beaker 2.}} \\
z_1 &= \mwl{pour(argmin(color(green),pos),pos(2))} \\
  x_2 &= \text{\nl{Then into the first beaker.}} \\
z_2 &= \mwl{actions[1](args[1][2],pos(3))}.
\end{align*}

We describe the notation we use for \reftab{features}, restricting
our discussion to actions that have two or fewer arguments. Our featurization
scheme, however, generalizes to an arbitrary number of arguments.
Given a logical form $z_i$, let $z_i.a$ be its action and
$(z_i.b_1, z_i.b_2)$ be its arguments (e.g., \wl{color(green)}).
The first and second arguments are
anchored over spans $[s_1,t_1]$ and $[s_2,t_2]$, respectively.
Each argument $z_i.b_j$ has a corresponding value $z_i.v_j$ (e.g., \wl{beaker1}),
obtained by executing $z_i.b_j$ on the context $c_i$. 
Finally, let $j,k \in \{1, 2\}$ be indices of the arguments. For example, we would
label the constituent parts of $z_1$ (defined above) as follows:
\begin{itemize}
  \setlength\itemsep{0pt}
\item $z_1.a = $ \mwl{pour}
\item $z_1.b_1 = $ \mwl{argmin(color(green),pos)}
\item $z_1.v_1 = $ \mwl{beaker3}
\item $z_1.b_2 = $ \mwl{pos(2)}
\item $z_1.v_2 = $ \mwl{beaker2}
\end{itemize}
 \section{Left-to-right parsing}

\begin{figure*}
\centering
\includegraphics[scale=0.40]{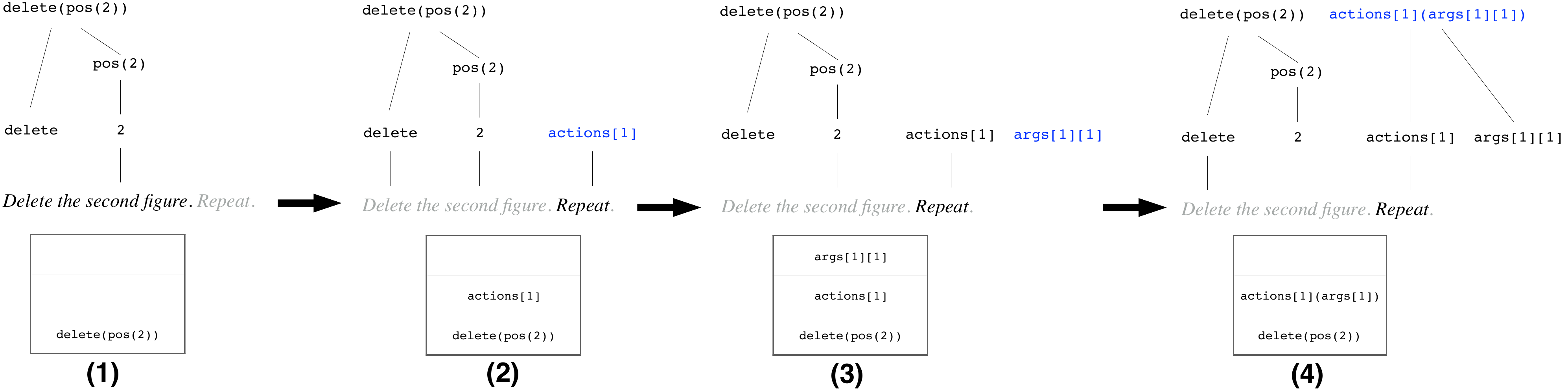}
\caption{
\label{fig:parsing}
  Suppose we have already constructed \wl{delete(pos(2))} for \nl{Delete the second figure.}
Continuing, we shift the utterance \nl{Repeat}.
Then, we build \wl{action[1]} aligned to the word \nl{Repeat.}
followed by \wl{args[1][1]}, which is unaligned.
Finally, we combine the two logical forms.
}
\end{figure*}

We describe a new parser suitable for learning from denotations in the
context-dependent setting.
Like a shift-reduce parser,
we proceed left to right,
but each \emph{shift} operation advances an entire utterance rather than one word.
We then sit on the utterance for a while, performing
a sequence of \emph{build} operations,
which either combine two logical forms on the stack (like the reduce operation)
or generate fresh logical forms,
similar to what is done in the floating parser of \citet{pasupat2015compositional}.

Our parser has two desirable properties:
First, proceeding left-to-right allows us
to build and score logical forms $z_i$ that depend on the world state $w_{i-1}$,
which is a function of the previous logical forms.
Note that $w_{i-1}$ is a random variable in our setting,
whereas it is fixed in \citet{zettlemoyer09context}.
Second, the \emph{build} operation allows us the flexibility
to handle ellipsis (e.g., \nl{Mix.}) and anaphora on full logical forms (e.g., \nl{Do it again.}),
where there's not a clear alignment between the words and the predicates generated.

The parser transitions through a sequence of hypotheses.
Each hypothesis is $h = (i, b, \sigma)$,
where $i$ is the index of the current utterance,
where $b$ is the number of predicates constructed on utterance $x_i$,
and $\sigma$ is a stack (list) of logical forms.
The stack includes both the previous logical forms $z_{1:i-1}$
and fragments of logical forms built on the current utterance.
When processing a particular hypothesis, the parser can choose
to perform either the shift or build operation:

\textbf{Shift:} The parser moves to the next utterance by
incrementing the utterance index $i$ and resetting $b$, which
transitions a hypothesis from $(i, b, \sigma)$ to $(i+1, 0, \sigma)$.

\textbf{Build:} The parser creates a new logical form by combining zero or
more logical forms on the stack.  There are four types of build operations:
\begin{enumerate}
  \item Create a predicate out of thin air (e.g., \wl{args[1][1]} in \reffig{parsing}).
    This is useful when the utterance does not explicitly reference
    the arguments or action. For example, in \reffig{parsing}, we are able to 
    generate the logical form \wl{args[1][1]} in the presence of ellipsis.
      \item Create a predicate anchored to some span of the utterance (e.g., \wl{actions[1]} anchored to \nl{Repeat}).
    This allows us to do credit assignment and capture which part of the utterance explains which part of the logical form.
  \item Pop $z$ from the stack $\sigma$ and push $z'$ onto $\sigma$,
  where $z'$ is created by applying a rule in \reftab{grammar} to $z$.
  \item Pop $z,z'$ from the stack $\sigma$ and push $z''$ onto $\sigma$,
    where $z''$ is created by applying a rule in \reftab{grammar} to $z,z'$
    (e.g., \wl{actions[1](args[1][1])} by the top-level root rule).
\end{enumerate}
The build step stops once a maximum number of predicates $B$ have been constructed
or when the top-level rule is applied.

We have so far described the search space over logical forms.
In practice, we keep a beam of the $K$ hypotheses with the highest score
under the current log-linear model.
 
\section{Model Projections}

Model A is ambitious, as it tries to learn from scratch how each word aligns to 
part of the logical form. For example, when Model A parses \nl{Mix it}, 
one derivation will correctly align \nl{Mix} to \wl{mix}, but others will
align \nl{Mix} to \wl{args[1][1]}, \nl{Mix} to \wl{pos(2)}, and so on (\reffig{relaxExample}).

As we do not assume a seed lexicon that could map \nl{Mix} to \wl{mix},
the set of anchored logical forms is exponentially large.
For example, parsing just the first sentence of \reffig{runningExample} would generate
1,216,140 intermediate anchored logical forms.

How can we reduce the search space?
The key is that the space of logical forms is \emph{much smaller} than the
space of anchored logical forms.
Even though both grow exponentially,
dealing directly with logical forms allows us to
generate \wl{pour} without the combinatorial choice over alignments.
We thus define Model B over the space of these logical forms.
\reffig{relaxExample} shows that the two anchored logical forms, which are treated differently
in Model A are collapsed in Model B. This dramatically reduces the search space; 
parsing the first sentence of \reffig{runningExample} generates 7,047 intermediate logical forms.

We can go further and notice that many compositional logical forms reduce to the same
flat logical form if we evaluate all the arguments.
For example, in \reffig{relaxExample},
\wl{mix(args[1][1])} and \wl{mix(pos(2))} are
equivalent to \wl{mix(beaker2)}.
We define Model C to be the space of these flat logical forms
which consist of a top-level action plus primitive arguments.
Using Model C, parsing the first sentence of \reffig{runningExample} generates
only 349 intermediate logical forms.

Note that in the context-dependent setting,
the number of flat logical
forms (Model C) still increases exponentially with the number of utterances,
but it is an overwhelming improvement over Model A.
Furthermore, unlike other forms of relaxation,
we are still generating logical forms that can express any denotation as before.
The gains from Model B to Model C hinge on the fact that in our world,
the number of denotations is much smaller than the number of logical forms.

\paragraph{Projecting the features.}

While we have defined the space over logical forms for Models B and C,
we still need to define a distribution over these spaces to to complete the picture.
To do this, we propose projecting the features of the log-linear model \refeqn{model}.
Define $\Piab$ to be a map from a anchored logical form $\za$ (e.g., \wl{mix(pos(2))} aligned to \nl{mix})
to an unanchored one $\zb$ (e.g., \wl{mix(pos(2))}),
and define $\Pibc$ to be a map from $\zb$ to the flat logical form $\zc$ (e.g., \wl{mix(beaker2)}).

We construct a log-linear model for Model B by constructing features $\phi(\zb)$
(omitting the dependence on $x_i,c_i$ for convenience)
based on the Model A features $\phi(\za)$.
Specifically, $\phi(\zb)$ is the component-wise maximum of $\phi(\za)$ over all
$\za$ that project down to $\zb$; $\phi(\zc)$ is defined similarly:
\begin{align}
  \phi(\zb) &\eqdef \max \{ \phi(\za) : \Piab(\za) = \zb \}, \\
  \phi(\zc) &\eqdef \max \{ \phi(\zb) : \Pibc(\zb) = \zc \}.
\end{align}
Concretely, Model B's features include indicator features over LF conditions in \reftab{features}
conjoined with every $n$-gram of the entire utterance, as there is no alignment.
This is similar to the model of \citet{pasupat2015compositional}.
Note that most of the derivation conditions (F2)--(F7) already depend on 
properties of the denotations of the arguments,
so in Model C, we can directly reason over the space of flat logical forms $\zc$ (e.g., \wl{mix(beaker2)})
rather than explicitly computing the max over more complex logical forms $\zb$ (e.g., \wl{mix(color(red))}).

\paragraph{Expressivity.}

In going from Model A to Model C,
we gain in computational efficiency, but we lose in modeling expressivity.
For example, for \nl{second green beaker} in \reffig{runningExample},
instead of predicting \wl{color(green)[2]}, we would have to predict
\wl{beaker3}, which is not easily explained by the words \nl{second green beaker}
using the simple features in \reftab{features}.

At the same time, we found that simple features can actually \emph{simulate} some logical forms.
For example, \wl{color(green)} can be explained by the feature that looks at
the \wl{color} property of \wl{beaker3}. Nailing \wl{color(green)[2]}, however, is not easy. 
Surprisingly, Model C can use a conjunction of features to express superlatives
(e.g., \wl{argmax(color(red),pos)}) by using one feature that places more mass 
on selecting objects that are red and another feature that places more mass 
on objects that have a greater position value.

 \section{Experiments}

Our experiments aim to explore the computation-expressivity tradeoff
in going from Model A to Model B to Model C.
We would expect that under the computational constraint of a finite beam size,
Model A will be hurt the most,
but with an infinite beam, Model A should perform better.

We evaluate all models on \emph{accuracy}, the fraction of examples that a model predicts correctly.
A predicted logical form $z$ is 
deemed to be correct for an example $(w_0,\bx,w_L)$ if the predicted logical form
$\bz$ executes to the correct final world state $w_L$.
We also measure the \emph{oracle accuracy}, which is the fraction of examples
where at least one $\bz$ on the beam executes to $w_L$. 
All experiments train for 6 iterations using AdaGrad \citep{duchi10adagrad}
and $L_1$ regularization with a coefficient of $0.001$.

\subsection{Real data experiments}

\paragraph{Setup.}
We use a beam size of 500 within each utterance, and prune to the top 5
between utterances.
For the first two iterations, Models B and C 
train on only the first utterance of each example ($L=1$).
In the remaining iterations, the models train on
two utterance examples.
We then evaluate on examples with $L=1, \dots, 5$,
which tests our models ability to extrapolate to longer texts.

\paragraph{Accuracy with finite beam.}

\begin{table}
  \small \centering
\begin{tabular}{llllllll}
  Dataset & Model & 3-acc & 3-ora & 5-acc & 5-ora\\ 
  \toprule
  \multirow{3}{*}{\alchemy} 
  & B & 0.189 & 0.258 & 0.037 & 0.055\\
  & C & \textbf{0.568} & \textbf{0.925} &\textbf{0.523} & \textbf{0.809}\\
  \midrule
  \multirow{3}{*}{\scene} 
  & B & 0.068 & 0.118 & 0.017 & 0.031\\ 
  & C & \textbf{0.232} & \textbf{0.431} &\textbf{0.147} &\textbf{0.253}\\
  \midrule
  \multirow{3}{*}{\tangrams} 
  & B & \textbf{0.649} & \textbf{0.910} & \textbf{0.276} & 0.513\\
  & C & 0.567 & 0.899 & 0.272 & \textbf{0.698}\\
  \bottomrule
\end{tabular}
\caption{Test set accuracy and oracle accuracy for examples containing $L=3$ and $L=5$ utterances.
Model C surpasses Model B in both accuracy and oracle on \alchemy{}  and \scene, whereas
Model B does better in \tangrams.}
\label{5_results}
\end{table}
 
We compare models B and C on the three real datasets 
for both $L=3$ and $L=5$ utterances (Model A was too expensive to use).
Table \ref{5_results} shows that
on 5 utterance examples, the flatter Model C achieves an average accuracy of 20\% higher than the
more compositional Model B.
Similarly, the average oracle accuracy is 39\% higher.
This suggests that
(i) the correct logical form often falls off the beam for Model B due to a larger search space,
and (ii) the expressivity of Model C is sufficient in many cases. 

On the other hand, Model B outperforms Model C on the \tangrams{} dataset. This happens for two reasons.
The \tangrams{} dataset has
the smallest search space, since all of the utterances refer to objects using position only. Additionally,
many utterances reference logical forms that Model C is unable to express, 
such as \nl{repeat the first step}, or \nl{add it back}.

\begin{figure*}
\subfloat[\alchemy]{\includegraphics[scale=0.30]{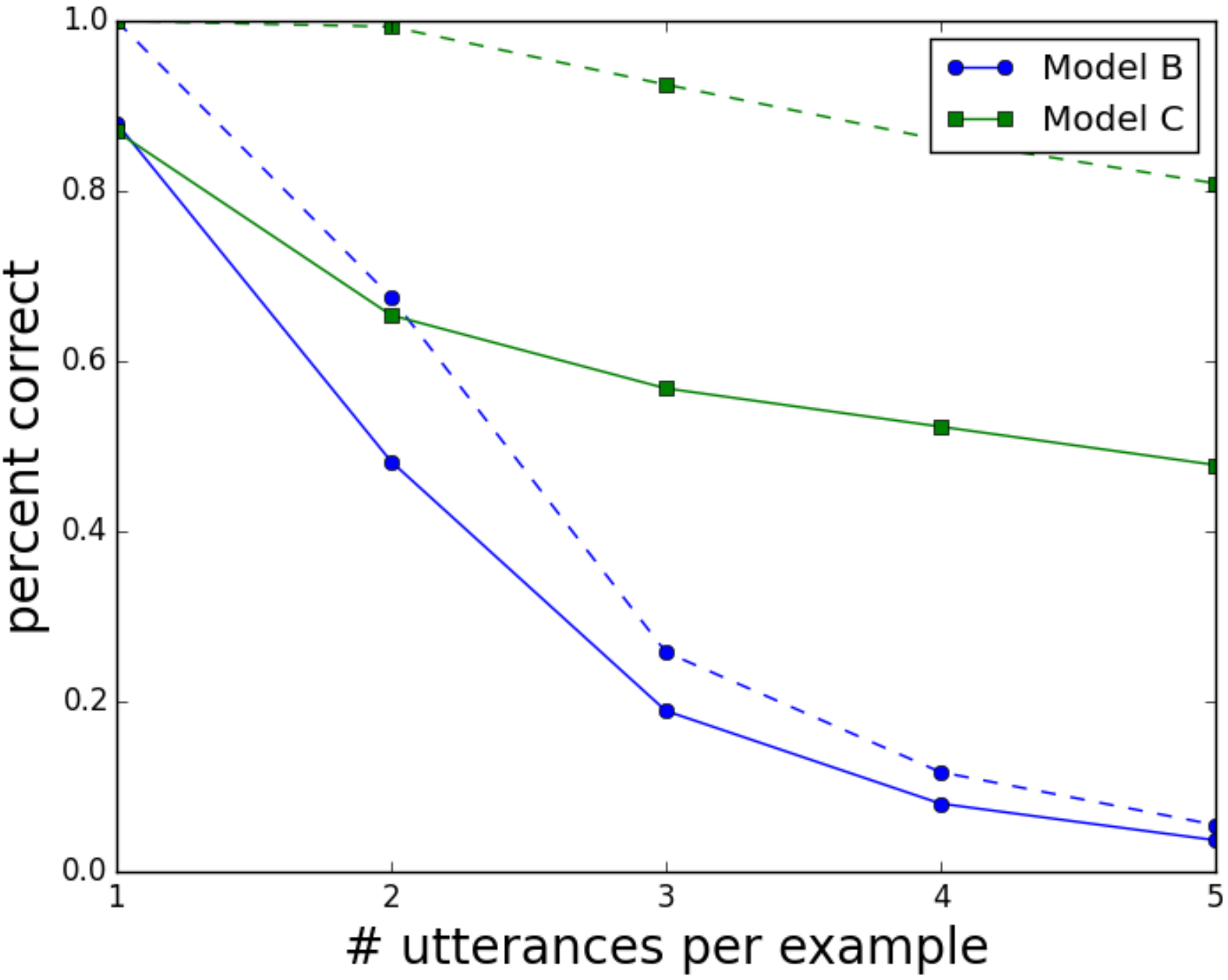}}\hspace{0.2em}
\subfloat[\scene]{\includegraphics[scale=0.30]{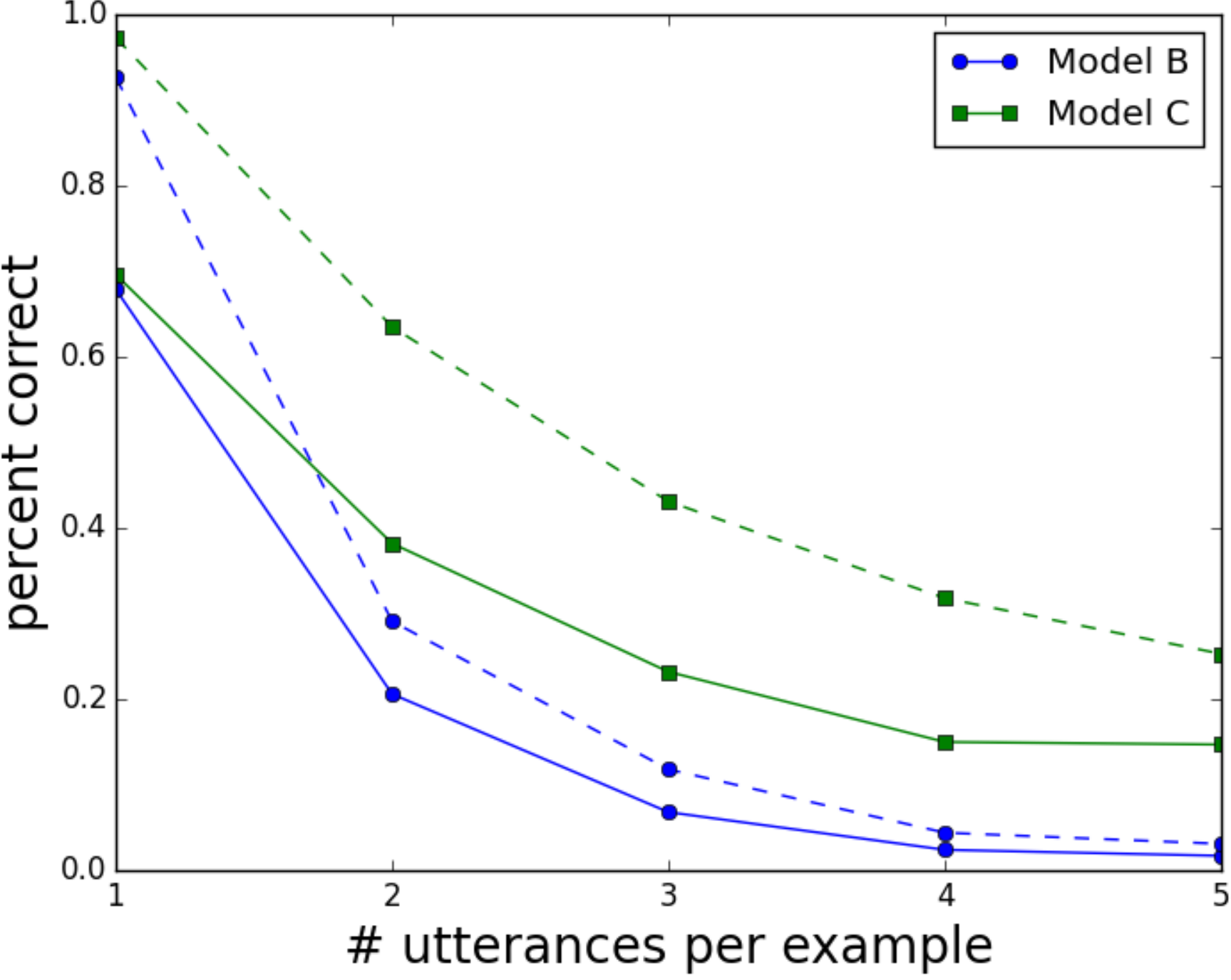}}\hspace{0.2em}
\subfloat[\tangrams]{\includegraphics[scale=0.30]{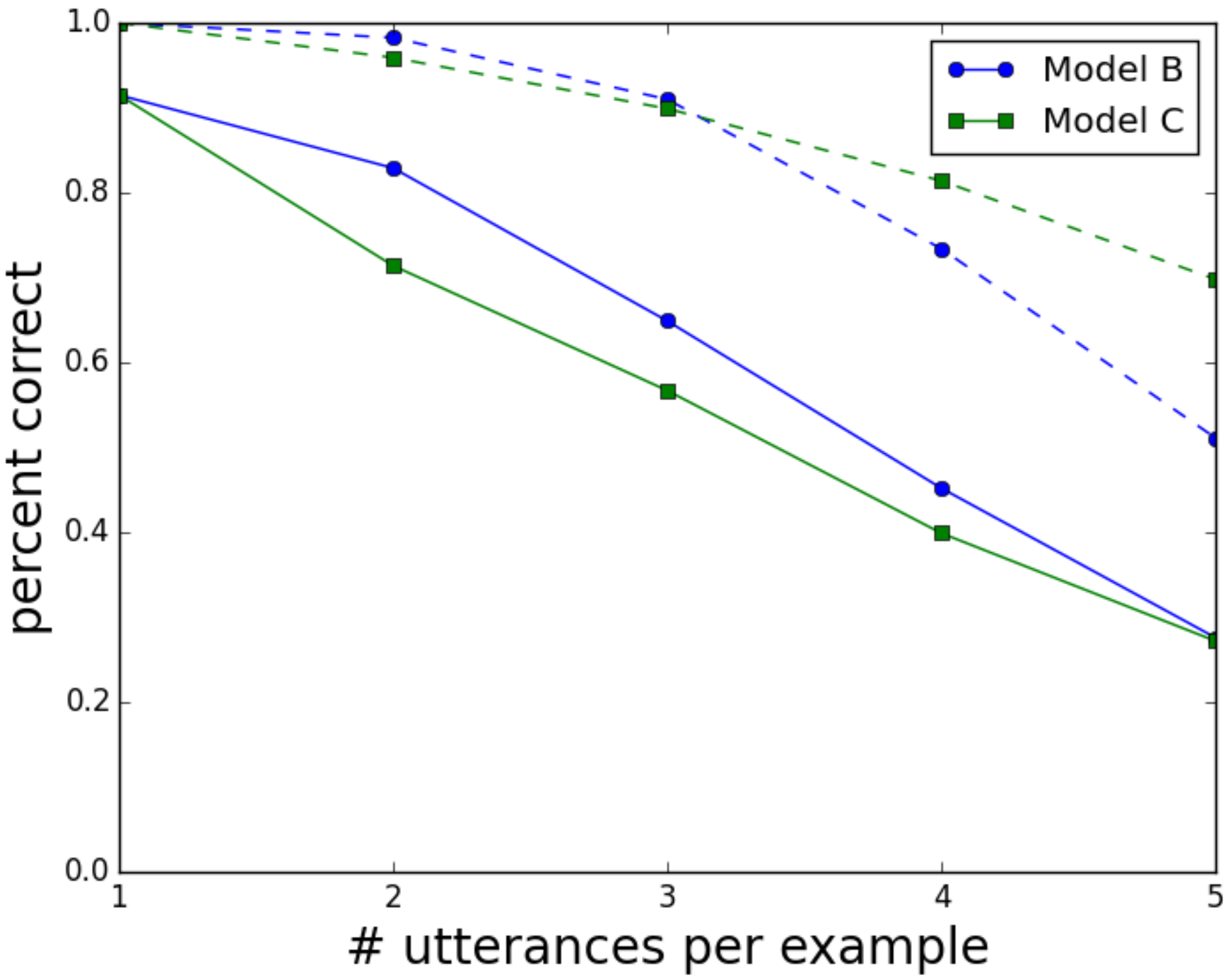}}\hspace{0.2em}
\caption{
Test results on our three datasets as we vary the number of utterances.
The solid lines are the accuracy, and the dashed line are the oracles:
With finite beam, Model C significantly outperforms Model B
on \alchemy{} and \scene{}, but is slightly worse on \tangrams{}.
}
\label{real_plot}
\end{figure*}

Figure \ref{real_plot} shows how the models perform as the number of utterances per
example varies. 
When the search space is small (fewer number of utterances),
Model B outperforms or is competitive with Model C. 
However, as the search space increases (tighter computational constraints), Model C does increasingly better. 

Overall, both models perform worse as $L$ increases,
since to predict the final world state $w_L$ correctly,
a model essentially needs to predict an entire sequence of logical forms $z_1, \dots, z_L$,
and errors cascade.
Furthermore, for larger $L$, the utterances tend to have richer context-dependence.

\subsection{Artificial data experiments}

\paragraph{Setup.}

Due to the large search space, running model A on real data is impractical. 
In order feasibly evaluate Model A,
we constructed an artificial dataset.
The worlds are created using the procedure described in \refsec{datasets}.
We use a simple template to generate utterances (e.g., \nl{drain 1 from the 2 green beaker}).

To reduce the search space for Model A,
we only allow actions (e.g., \wl{drain}) to align to verbs
and property values (e.g., \wl{green}) to align to adjectives.
Using these linguistic constraints provides a slightly
optimistic assessment of Model A's performance.

We train on a dataset of 500 training examples and evaluate on 500 test examples.
We repeat this procedure for varying beam sizes, from $40$ to $260$.
The model only uses features (F1) through (F3).

\paragraph{Accuracy under infinite beam.}

\begin{figure}[h]
\begin{center}
\includegraphics[scale=0.40]{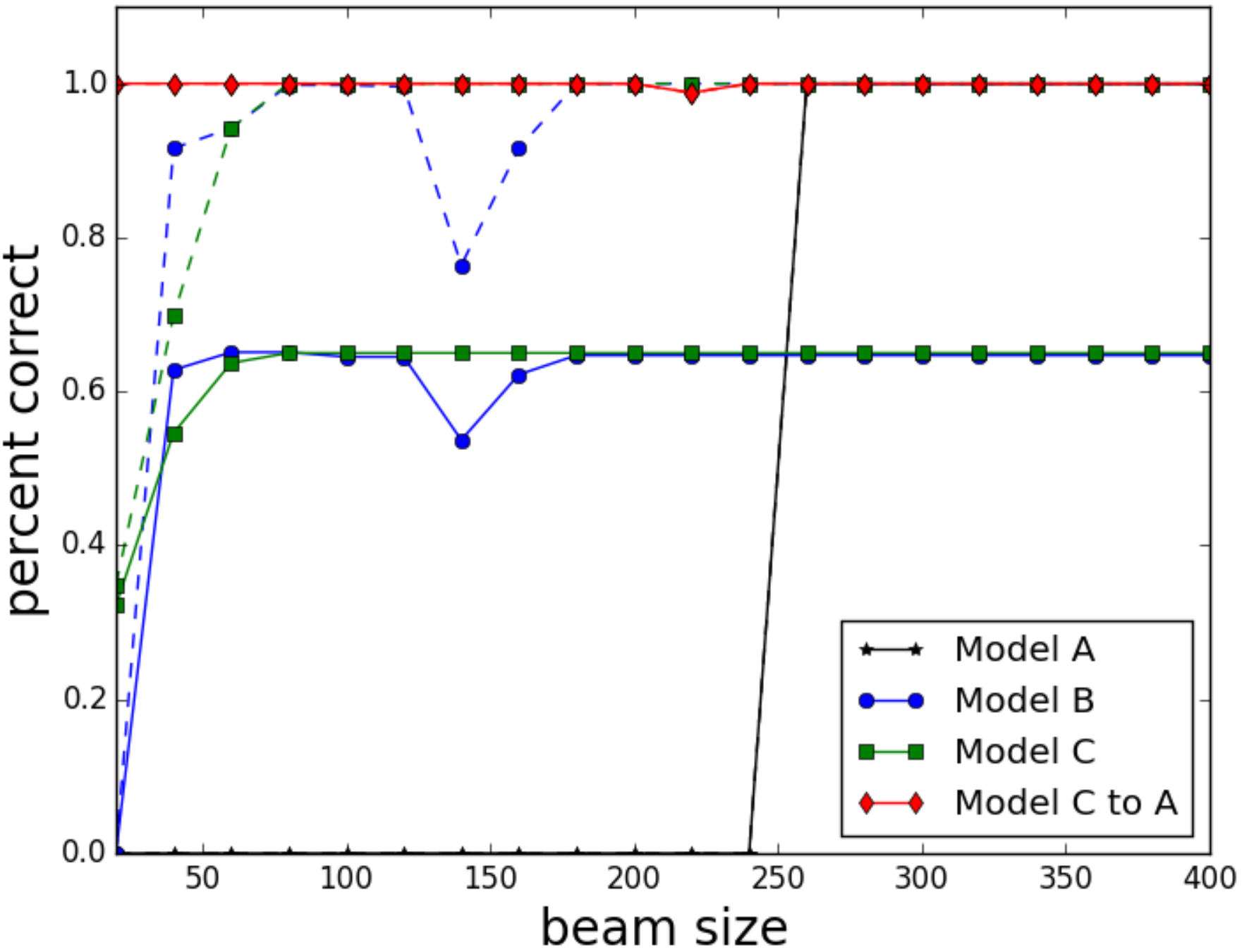}
\end{center}
\caption{Test results on our artificial dataset with varying beam sizes.
  The solid lines are the accuracies,
  and the dashed line are the oracle accuracies.
  Model A is unable to learn anything with beam size $<240$.
  However, for beam sizes larger than $240$, Model A attains
  $100\%$ accuracy. Model C does better than Models A and B
  when the beam size is small $<40$, but otherwise 
  performs comparably to Model B. Bootstrapping Model A using
  Model C parameters outperforms all of the other models and
  attains $100\%$ even with smaller beams. }
\label{fig:bootstrap}
\end{figure}

Since Model A is more expressive, we would expect it to be more powerful
when we have no computational constraints.
\reffig{bootstrap} shows that this is indeed the case:
When the beam size is greater than 250, all models attain an oracle of 1,
and Model A outperforms Model B, which performs similarly to Model C.
This is because the
alignments provide a powerful signal for constructing the logical forms.
Without alignments, Models B and C learn noisier features, and accuracy
suffers accordingly. 

\paragraph{Bootstrapping.}
Model A performs the best with unconstrained computation, and Model C performs 
the best with constrained computation. Is there some way to bridge the two? 
Even though Model C has limited expressivity, it can still learn to associate
words like \nl{green} with their corresponding predicate \wl{green}.
These should be useful for Model A too.

To operationalize this,
we first train Model C and use the parameters to initialize model A.
Then we train Model A.
\reffig{bootstrap} shows that although Model A and C predict different logical forms,
the initialization allows Model C to A to perform well 
in constrained beam settings.
This bootstrapping works here because Model C is a \emph{projection} of Model A,
and thus they share the same features.

\subsection{Error Analysis}

\begin{figure}
\centering
\includegraphics[scale=0.60]{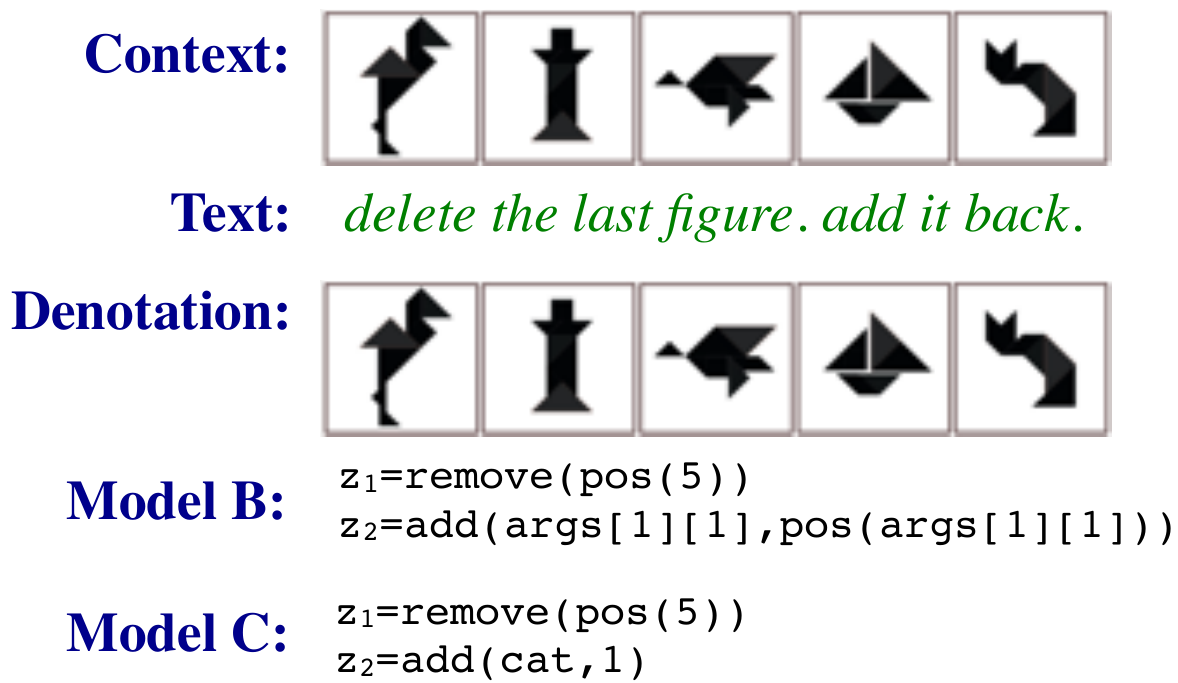}
\caption{Predicted logical forms for this text:
The logical form \wl{add} takes a figure and
position as input. Model B predicts the correct logical form.
Model C does not understand that \nl{back} refers
to position 5, and adds the cat figure to position 1.
}
\label{fig:anecdote}
\end{figure}

\begin{table}[t]
\small \centering
\begin{tabular}{llllll}
model & beam & action & argument & context & noise \\ \toprule
B     & 0.47 & 0.03   & 0.17     & 0.23    & 0.04  \\ 
C     & 0.15 & 0.03   & 0.25     & 0.5    & 0.07  \\ \bottomrule
\end{tabular}
\caption{Percentage of errors for Model B and C:
Model B suffers predominantly from computation constraints,
while Model C suffers predominantly from a lack of expressivity.
}
\label{tab:errors}
\end{table}

We randomly sampled 20 incorrect predictions on 3 utterance examples from each
of the three real datasets for Model B and Model C.
We categorized each prediction error into one of the following categories:
(i) logical forms falling off the beam;
(ii) choosing the wrong action (e.g., mapping \nl{drain} to \wl{pour});
(iii) choosing the wrong argument due to misunderstanding the description
(e.g., mapping \nl{third beaker} to \wl{pos(1)});
(iv) choosing the wrong action or argument due to misunderstanding of context (see \reffig{anecdote});
(v) noise in the dataset.
\reftab{errors} shows the fraction of each error category.
 \section{Related Work and Discussion}

\paragraph{Context-dependent semantic parsing.}

Utterances can depend on either linguistic context or world state context.
\citet{zettlemoyer09context} 
developed a model that handles references to previous logical forms;
\citet{artzi2013weakly}
developed a model that handles references to the current world state.
Our system considers both types of context,
handling linguistic phenomena such as ellipsis and anaphora
that reference both previous world states and logical forms.

\paragraph{Logical form generation.}

Traditional semantic parsers generate logical forms by
aligning each part of the logical form to the utterance 
\citep{zelle96geoquery,wong07synchronous,zettlemoyer07relaxed,kwiatkowski11lex}.
In general, such systems rely on a lexicon,
which can be hand-engineered,
extracted \citep{cai2013large,berant2013freebase},
or automatically learned from annotated logical forms
\citep{kwiatkowski10ccg,chen12lexicon}. 

Recent work on learning from denotations
has moved away from anchored logical forms.
\citet{pasupat2014extraction} and \citet{wang2015overnight}
proposed generating logical forms without alignments,
similar to our Model B.
\citet{yao2014freebase} and \citet{bordes2014qa} have 
explored predicting paths in a knowledge graph directly,
which is similar to the flat logical forms of Model C. 

\paragraph{Relaxation and bootstrapping.}

The idea of first training a simpler model in order to work up to a more
complex one has been explored other contexts.
In the unsupervised learning of generative models,
bootstrapping can help escape local optima and provide helpful regularization \citep{och03systematic,liang09semantics}.
When it is difficult to even find one logical form that reaches the denotation,
one can use the relaxation technique of \citet{steinhardt2015relaxed}.

Recall that projecting from Model A to C creates a more computationally
tractable model at the cost of expressivity.
However, this is because Model C used a linear model.
One might imagine that a non-linear model
would be able to recuperate some of the loss of expressivity.
Indeed, \citet{neelakantan2016neural} use recurrent neural networks
attempt to perform logical operations.
One could go one step further and 
bypass logical forms altogether,
performing all the logical reasoning in a continuous space
\citep{bowman2014recursive,weston2015towards,guu2015traversing,reed2016neural}.
This certainly avoids the combinatorial explosion of logical forms in Model A,
but could also present additional optimization challenges.
It would be worth exploring this avenue
to completely understand the computation-expressivity tradeoff.

\section*{Reproducibility} Our code, data, and experiments are available
on CodaLab at {\small \url{https://worksheets.codalab.org/worksheets/0xad3fc9f52f514e849b282a105b1e3f02/}}.
\section*{Acknowledgments}
We thank the anonymous reviewers for their constructive feedback. 
The third author is supported by a Microsoft Research Faculty
Fellowship.

\bibliographystyle{acl2016}
\bibliography{refdb/all}

\end{document}